\pdfoutput=1

\documentclass[11pt]{article}

\usepackage[preprint]{neurips_2022}

\usepackage{times}
\usepackage{latexsym}
\usepackage[T1]{fontenc}

\usepackage[utf8]{inputenc}

\usepackage{microtype}
\usepackage{amsfonts}
\usepackage{bbding}

\usepackage{inconsolata}
\usepackage{hyperref}

\usepackage{algorithm}
\usepackage{algpseudocode}

\usepackage{subcaption}
\usepackage{graphicx}
\usepackage{amsmath}
\usepackage{multirow}
\usepackage{caption}
\usepackage{epsfig}
\usepackage{xcolor}
\usepackage{CJKutf8}
\usepackage{enumitem}
\usepackage{booktabs}

\title{Automatic Prompt Selection for Large Language Models}

\author{Viet-Tung Do$^{1}$, Van-Khanh Hoang$^{1}$, Duy-Hung Nguyen$^1$, Shahab Sabahi$^1$,\\ \textbf{Jeff Yang$^1$, Hajime Hotta$^1$, Minh-Tien Nguyen$^{1,2}$ and Hung Le$^3$} \\
$^1$ Cinnamon AI, Vietnam.\\
\texttt{\{ace, kay, hector, sshahab, jeff.yang, hajime, ryan.nguyen\}@cinnamon.is} \\
$^2$ Hung Yen University of Technology and Education, Vietnam.\\
\texttt{tiennm@utehy.edu.vn}\\
$^3$ Deakin University, Australia.\\
\texttt{thai.le@deakin.edu.au}
}

\begin{document}
\maketitle
\begin{abstract}

Large Language Models (LLMs) can perform various natural language processing tasks with suitable instruction prompts. However, designing effective prompts manually is challenging and time-consuming. Existing methods for automatic prompt optimization either lack flexibility or efficiency. In this paper, we propose an effective approach to automatically select the optimal prompt for a given input from a finite set of synthetic candidate prompts. Our approach consists of three steps: (1) clustering the training data and generating candidate prompts for each cluster using an LLM-based prompt generator; (2) synthesizing a dataset of input-prompt-output tuples for training a prompt evaluator to rank the prompts based on their relevance to the input; (3) using the prompt evaluator to select the best prompt for a new input at test time. Our approach balances prompt generality-specificity and eliminates the need for resource-intensive training and inference. It demonstrates competitive performance on zero-shot question-answering datasets: GSM8K, MultiArith, and AQuA.

\end{abstract}

\section{Introduction}
Large Language Models (LLMs) have emerged as a cornerstone in natural language processing (NLP), propelled by advancements in scaling techniques and attention mechanisms. These colossal models have a remarkable ability to generate texts based on initial inputs, showcasing extraordinary proficiency across diverse tasks, particularly in zero-shot and few-shot scenarios \cite{brown2020language, srivastava2023beyond, nguyen2023giant}. The key to unlocking this exceptional performance heavily depends on the choice of prompts--extra text added to the initial input to query LLMs. Recent studies underscore the pivotal role of well-crafted prompts in enhancing the accuracy, adequacy, and relevance of LLM responses across various tasks such as question answering, text summarization, and text generation \cite{lu2022fantastically, lyu2023llm, white2023prompt}. Effective prompting reduces the necessity for resource-intensive fine-tuning on task-specific data, offering a cost-effective and time-saving alternative. Additionally, when provided with appropriate prompts, LLMs can engage in intricate reasoning and problem-solving. This can be achieved by breaking down complex problems into manageable steps or by providing illustrative examples and demonstrations \cite{yao2022react, wei2022chain}. 

Prompting provides a good method for working with LLMs for many NLP tasks, yet, we argue that manual prompting is a nontrivial task. It requires careful engineering and testing to achieve the best results, which is time-consuming and labor-expensive. Early attempts for automatic prompting use gradient descent optimization on prompt token embeddings, a.k.a "soft prompts" \cite{qin2021learning}. Alternatively, prompt generation directly generates prompt tokens, which enhance human interpretability and flexibility \cite{deng2022rlprompt,zhang2022tempera}. In this direction, LLMs are viewed as frozen black boxes where different optimization methods such as evolutionary algorithm (EA) and reinforcement learning (RL) can be applied to search for an optimal prompt for all inputs of the downstream task \cite{prasad2023grips,deng2022rlprompt}. Other methods use RL to dynamically generate specific prompts for each input \cite{zhang2022tempera}. On a different front, LLM-based approaches employ multiple LLMs to generate and explore better prompts \cite{zhou2022large, yang2023large,liao2022zero}.

One limitation of current approaches is that they struggle to strike a balance between prompt generality and specificity, either relying on a single prompt for all inputs (lacking flexibility) or generating a distinctive prompt per input (expanding the search space and potentially destabilizing the system).  Intuitively, prompts should change only when there is a shift in input type. For instance, mathematical data featuring algebraic inputs necessitates an algebra-related prompt, while geometry inputs require a geometry-related prompt. Also, current methods are inefficient in computation. EA and RL-based approaches involve complex optimization and numerous iterations with downstream LLMs in training. Furthermore, LLM-based methods need to query multiple LLMs during training or inference, which is slow and expensive.

In this paper, we address the identified limitations of discrete prompt optimization. Our approach combines prompt generation and ranking to find the best prompt for a given input and task, achieving a balance between prompt generality and specificity while ensuring computational efficiency. Our method, dubbed Automatic Prompt Selection (APS), consists of three steps: (1) we cluster the task data into coherent groups and use LLM-based prompt generators to create a set of prompts for each group. We combine these sets to form a prompt database that covers different input types. (2) We use the prompts in the database to query an LLM and generate a dataset of tuples, each containing an input, a prompt, an LLM output, and a ground-truth output. We train a prompt evaluator on this dataset using a preference loss, which assigns higher scores to prompts that produce outputs closer to the ground truth. (3) Given an input, we "generate" an optimal prompt by selecting the one from the prompt database with the highest score, as determined by the prompt evaluator. In short, we summarize our contributions as follows.
\begin{itemize}
\item We propose a novel approach for discrete prompt optimization, aiming to identify the optimal prompt from a finite fixed set generated by another LLM. This strategy minimizes the search space while maintaining a degree of flexibility to strike a balance between prompt generality and specificity. Our approach is lightweight, necessitating only standard supervised training on offline data and enabling cost-effective inference without additional LLM access.
\item We conduct extensive experiments on three challenging QA datasets and show that our method outperforms state-of-the-art baselines
\end{itemize}


\section{Related Works}
Prompt engineering plays a crucial role in optimizing LLMs for various tasks, whether performed manually or automatically \cite{gao2021making, reynolds2021prompt}. This field encompasses different strategies, such as prompt tuning \cite{qin2021learning}, prompt generation \cite{zhang2022tempera, singh2022explaining}, and prompt selection for specific tasks \cite{liao2022zero, sorensen2022information, zhang2022automatic}. Prompt tuning methods employ gradient-based searches to refine prompts \cite{qin2021learning, pang-etal-2023-sharpt}. However, these methods typically require access to the LLM's parameters, limiting their applicability to LLMs with restricted access to training data. In contrast, our approach belongs to the class of prompt generation, treating downstream LLMs as black boxes and seeking optimal token-level prompts without fine-tuning token embeddings.
\begin{figure*}[t]
  \caption{Automatic Prompt Selection (APS) has three steps. \textit{(1) Prompt Database Generation:} We cluster training data, use LLM-based prompt generator $\mathtt{A}_1$ for diverse prompts in each group, and combine them into a versatile prompt database. \textit{(2) Prompt Evaluator Training:} We query a data generation LLM $\mathtt{A}_2$ with generated prompts and training inputs to generate tuples: input ($q,c$), prompt ($p$), LLM output ($a'$), and ground-truth output ($a'$). Training the prompt evaluator $\mathtt{E}$ on this dataset, we adopt a preference loss, encouraging high scores for good prompts and low scores for bad ones. \textit{(3) Prompt Ranking:} During inference, given a testing input, we pick the highest-scoring prompt from the database with the help of the prompt evaluator. The selected prompt (highlighted with red border) will be used with a downstream LLM $\mathtt{M}$ to compute the final output $a'$.}\label{img:model}
  \centering
\includegraphics[width=1\textwidth]{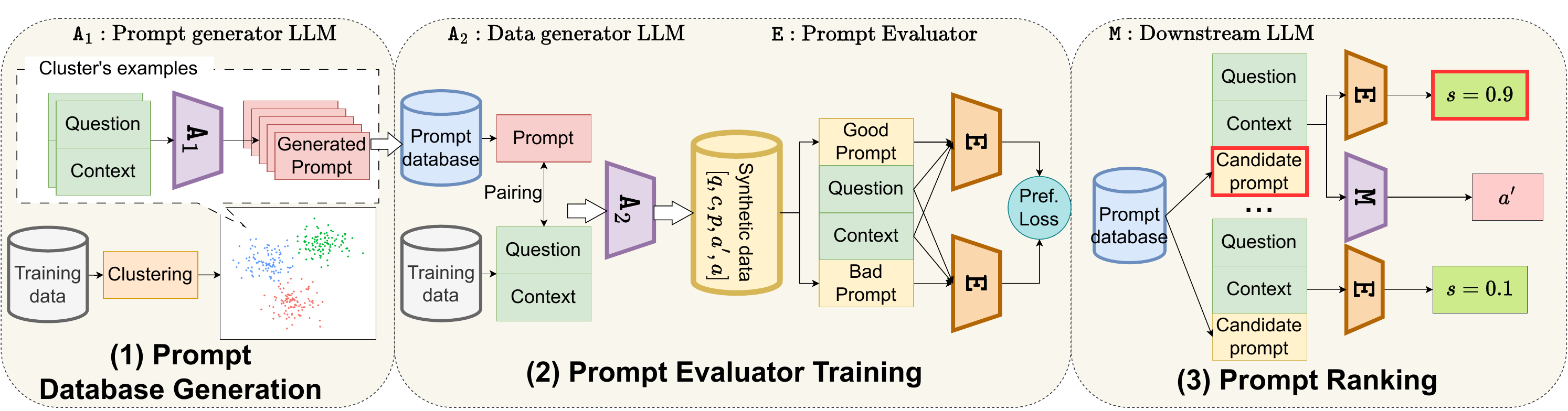}
\end{figure*}

Prompt generation methods leverage different optimization techniques to directly generate prompt tokens. For example, RL-Prompt \cite{deng2022rlprompt} and TEMPERA \cite{zhang2022tempera} employ reinforcement learning, while GRIPS \cite{prasad2023grips} relies on evolutionary algorithms. These approaches involve complex optimization processes with extensive interactions with LLMs during training. In contrast, our method only requires standard supervised training using offline data, making it efficient and cost-effective.
Another line of research involves leveraging the in-context learning capability of LLMs as prompt optimizers. For instance, 
APE \cite{zhou2022large} uses LLMs to generate multiple instruction candidates, evaluates them using the target model, and selects the most effective instruction based on calculated evaluation scores. OPRO \cite{yang2023large} iteratively generates and evaluates multiple prompts, selecting the most effective ones based on evaluation scores. These approaches apply the generated prompts to all inputs.  In contrast, our approach focuses on generating diverse prompts and selecting the topmost suitable ones for different inputs. This strategy strikes a balance between exploration and exploitation–a dynamic adaptation to contexts-enhancing the effectiveness of LLMs across diverse applications.

Prompt selection methods, such as ZPS \cite{liao2022zero}, MI \cite{sorensen2022information}, and Auto-CoT \cite{zhang2022automatic}, aim to identify high-quality prompts tailored to specific tasks and inputs. ZPS filters out underperforming prompts and assembles the remaining prompts for data labeling, while MI maximizes mutual information between inputs and model predictions. We note that these methods require an initial human-crafted prompt database or templates whereas we automatically generate them. Auto-CoT clusters and generates reasoning chains for questions, selecting representative questions from each cluster to create a diverse set of examples. However, these prompt selection methods typically require LLMs to produce outputs for all candidate prompts before ranking them, leading to significant computational costs or latency. To address this challenge, our method introduces a trainable prompt evaluator that is capable of ranking prompts in our database without executing the LLMs. This approach significantly reduces computational overhead and latency while maintaining the robustness and diversity of the prompts.
\section{Automatic Prompt Selection}
\subsection{Problem Statement}
The task is to find an optimal prompt for a given question and its relevant context.
Formally, we define a prompt $p$ as an instruction added to the original input.  A downstream task LLM $\mathtt{M}$ will take the prompt-augmented input and generate an output. In the question answering (QA) setting, the input can consist of a question $q$ and an optional context $c$, thus the generated output can be represented as $\mathtt{M}(q,c,p)$.  Our objective is to find an optimal prompt generator $\mathtt{G}^*$ so that for every question and context $(q,c)$, we can sample a prompt $p$ from the prompt generator to guide the LLM $\mathtt{M}$ to produce an output matching the correct output $a$. Formally, the selection can be shown as follows.
\begin{equation}
    \mathtt{G}^* = \arg \max_{\mathtt{G}} \mathbb{E}_{(a,q,c),p \sim \mathtt{G}(q,c)} f(\mathtt{M}(q, c, p), a)
\end{equation}
where $f(\cdot)$ can be any evaluating functions such as exact matching $1_{\mathtt{M}(q, c, p) = a}$. We note that the training data is only used for learning the prompt generator. The LLM $\mathtt{M}$ is not fine-tuned or does not access the training data, thus, our downstream task is equivalent to zero-shot QA. 


This task becomes notably challenging when employing traditional prompt generation approaches that utilize generative models to craft a new prompt for each sample, leading to extensive prompt search space. The process turns into cost-prohibitive with multiple iterations of querying LLMs using these new prompts. To significantly lower the cost of prompting, we suggest replacing the generative model with a prompt evaluator. This evaluator can assign a score to a tuple $(q,c,p)$, indicating the fitness of the prompt $p$ for a question-context input. We then iterate over a fixed representative prompt database to choose the prompt with the highest score given by the evaluator. In the next sections, we will detail  3 steps of our approach: prompt database generation, prompt evaluator training, and prompt ranking (see Figure \ref{img:model}).

\subsection{Prompt Database Generation}
\label{subsec:prompt_database}
 We argue that similar inputs (question-context pairs) benefit equally from the same prompt.
 This is because similar inputs share a common meaning represented in the form of tokens.
 We therefore create a prompt database used for matching with the inputs.
 The database generation consists of two steps: clustering and meta-prompt generation.

 \paragraph{Clustering}
 The first step is to assign the training data $\mathcal{D}_{train}$ into different coherent groups (clusters), expecting that inputs from the same group can share the same prompt. Specifically, we use Sentence-Transformer \cite{reimers-2019-sentence-bert} to encode the concatenations of the input questions and the corresponding contexts which are then fed into $K$-Means clustering algorithm \cite{macqueen1967some} to calculate the clusters.

\paragraph{Prompt generation}
Once clusters have been formed, the second step is to generate prompts for each cluster. The generation follows the generative approach of APE (Automatic Prompt Engineer) \cite{zhou2022large}. Specifically, we create a meta-prompt consisting of a few question-context-answer tuples from each cluster $(q,c,a)$ as demonstrative examples and an instruction to ask an additional LLM $\mathtt{A}_1$ to generate candidate prompts.
We conduct a basic exact match checking to eliminate duplicate prompts, ensuring that the prompt database consists of distinct entries.
Compared to manual prompts from humans with limited patterns, this approach provides more freedom to creatively produce diverse candidate prompts. The union of all generated prompts forms the prompt database. We can control the size of the prompt database by specifying hyperparameters such as $c$- the number of clusters, and $n_p$- the number of generated prompts per cluster. The prompt database generation process is summarized in Algorithm \ref{algo:prompt_database}.


\begin{algorithm}
\caption{Prompt database generation.}
\label{algo:prompt_database}
\begin{algorithmic}[1]
\Require Training data $\mathcal{D}_{train}$, clustering algorithm $C$, LLM-based prompt generator $\mathtt{A}_1$, hyperparameters $c$ and $n_p$.
\State Initialize the prompt database $\mathcal{P} = \emptyset $
\State Group the training data into $c$ clusters $\mathcal{G}_1, \mathcal{G}_2, .., \mathcal{G}_{c} = C(\mathcal{D}_{train})$ 
\For{$\mathcal{G}$ in $\mathcal{G}_1, \mathcal{G}_2, .., \mathcal{G}_{c}$}
    \State Randomly construct a set of demonstrations $\mathcal{D}_{demo} \subset \mathcal{G}$ to build meta-prompt $p_{meta}$
    \State Generate prompts $\mathcal{P}'=\mathtt{A}_1(p_{meta},n_p)$
    \State $\mathcal{P} \gets \mathcal{P} \cup \mathcal{P}'$
\EndFor
\State \Return Prompt database $\mathcal{P}$
\end{algorithmic}
\end{algorithm}


\subsection{Prompt Evaluator Training}
Our objective is to develop a model capable of evaluating the relevance of a prompt by assigning a corresponding score to it. The model called prompt evaluator ($\mathtt{E}$), receives input question $q$, a context text $c$, and a prompt $p$ to produce a score that gauges the relevance of the prompt to the input. 
\begin{equation}
s=\mathtt{E}_{\theta}(q,c,p)    
\end{equation}

To find the parameter $\theta$ of the prompt evaluator, we adopt a preference learning scheme by preparing a comparison dataset \cite{christiano2017deep}. With this setting, we can still estimate the score for each prompt following the Bradley-Terry model without requiring the exact score labels \cite{bradley1952rank}.
The training process is done in two steps: data collection and evaluator training.

\paragraph{Data collection}
The first step of training the evaluator is to collect training samples. To do that, given a separate question-context pair $(q, c)$ and the correct answer $a$ belonging to a group $\mathcal{G}$, we partition the set of prompts of that group $\mathcal{P}_{\mathcal{G}}$ into two subsets called good prompts ($\mathcal{P}_{q, c}^{good}$) and bad prompts ($\mathcal{P}_{q, c}^{bad}$).
A good prompt is a prompt, combined with the input question and its context, that can be fed into an LLM $\mathtt{A}_2$ to output an answer that is very similar to the correct answer $a$ in terms of meaning.
On the other hand, a bad prompt tends to make the LLM $\mathtt{A}_2$ to produce a different answer compared to the correct answer $a$. The collection of good and bad prompts is shown as follows.
\begin{equation}
    \centering
    \mathcal{P}_{q, c}^{good} = \{f(\mathtt{A}_2(q, c, p), a) \geq \lambda, \forall p \in \mathcal{P}_\mathcal{G} \}
\end{equation}
\begin{equation}
    \mathcal{P}_{q, c}^{bad} = \{f(\mathtt{A}_2(q, c, p), a) < \lambda, \forall p \in \mathcal{P}_\mathcal{G} \}
\end{equation}
where $\lambda = (\max f - \min f) / 2$ is a decision threshold. Notably, we only collect the good and bad prompts for only $m$ closest question-context pairs to the centroid of each cluster to ignore unrepresentative inputs and ensure a manageable synthetic data size. We note that the data collection is done automatically without human evaluation.



\paragraph{Evaluator training}
Once the two distinct prompt sets have been created, we form all possible bi-partite combinations as the training data for the prompt evaluator.
With each sample $(q, c)$ with the correct answer $a$, the evaluator samples a good prompt $p_{good}$ and a bad prompt $p_{bad}$ so that the tuple of $(q, c, p_{good})$ can output an answer $a'$ that approximates the correct answer $a$ and the tuple of $(q, c, p_{bad})$ outputs a different answer compared to the correct answer $a$. In this sense, $\mathtt{E}()$ returns a high score if it receives a good prompt while outputting a low score with a bad prompt.
By doing that, we force the evaluator to select a good prompt rather than a bad prompt.
The prompt evaluator $\mathtt{E}$ is trained by minimizing the preference loss:
\begin{equation}
    \mathcal{L} = -\sum_{\mathcal{P}_{q, c}^{good}} \sum_{\mathcal{P}_{q, c}^{bad}} \log (\mathtt{E}_\theta (q, c, p_{good}) - \mathtt{E}_\theta (q, c, p_{bad}) + \epsilon) 
\end{equation}
With the margin $\epsilon$, we can control the minimum distance between a correct input and an incorrect one to reduce noisy model predictions.


\subsection{Prompt Ranking}\label{sec:ranking}
At test time, given a new input $(q', c')$, we use the prompt evaluator $\mathtt{E}$ to calculate the score for all candidate prompts in the prompt database $\mathcal{P}$ when pairing them with the input. Finally, the selected prompt $p^*$ is the one with the highest score. Hence, the prompt generator can be formulated as:
\begin{equation}
\label{eq:ranking}
p^*=\mathtt{G}(q',c') = \arg \max_{p\in \mathcal{P}} \mathtt{E}_\theta(q',c', p)
\end{equation}
A notable benefit of our approach is the ability to rank all prompts in the databases and choose the top-$k$ best prompts. These prompts can produce varied responses, and we can utilize voting mechanisms to ascertain the most common outputs generated by the majority of the prompts: 
\begin{equation}
    a' = \mathtt{Mode} (\{\mathtt{M}(q, c, p^*_i), \forall i=1 \ldots k\})
\end{equation}
where $\{p^*_i\}_{i=1}^k$ denotes prompts with the top-$k$ scores defined in Eq. \eqref{eq:ranking}.
The voting mechanism acts as a safeguard against potential biases or outliers associated with specific prompts, ensuring a more balanced and accurate result.






\section{Experimental Setup}
\subsection{Datasets}
We examine the proposed method on challenging question-answering datasets that necessitate non-trivial prompts. Specifically, we concentrate on arithmetic problems because of their intricate reasoning processes. Unlike typical retrieval question-answering scenarios, extractive approaches cannot be employed to solve arithmetic questions. We used three benchmark datasets for evaluation. Table \ref{tab:dataset} summarizes datasets' information.

\paragraph{GSM8K} consists of 8.5K high-quality linguistically diverse grade school math word problems created by human problem writers \cite{karl2021gsm8k}. The dataset is segmented into 7.5K training and 1K test problems. These problems require between 2 and 8 steps to solve, and solutions primarily involve performing a sequence of elementary calculations using basic arithmetic operations $\pm\times\div$ to reach the final answer.

\paragraph{MultiArith} \cite{roy2016multiarith} is the subset of the Mawps repository \cite{rik2016mathworld}. This dataset is a collection of mathematical problems that are specifically designed to test the ability of machine learning models to perform complex arithmetic operations and reasoning. These problems demand the application of multiple arithmetic operations and logical reasoning to be successfully solved.

\paragraph{AQuA} consists of more than 100,000 algebraic word problems that include both answers and natural language rationales \cite{ling2017aqua}. Unlike other datasets, here, each data sample contains four parts: question, options (context), rationale, and the correct option. We measure accuracy on the correct option, and only use the rationale for the clustering step. 


\begin{table}[!t]
    \centering
    \caption{Three QA arithmetic datasets. FF and MC mean free form and multiple choice, respectively.} \label{tab:dataset}
    \begin{tabular}{cccc} \hline
       Dataset  & Train & Test & Answer type \\ \hline
        GSM8K  & 7473 & 1319 & FF\\
        MultiArith & 420 & 180 & FF\\
        AQuA & 97467 & 254 & MC\\
        \hline
    \end{tabular}
\end{table}
\subsection{LLM Models}
We used the GPT-3.5-turbo model for both the prompt generator $\mathtt{A}_1$, the data synthesizer $\mathtt{A}_2$, and the downstream task solver $\mathtt{M}$. To maintain consistent and replicable results, we created and used only one API version and model deployment name. The model name is gpt-3.5-turbo-0613 with the version of 2023-05-15. We set the temperature as 0.0 to restrict the randomness of GPT when decoding. The top $p$ parameter is set as 1.0 and the maximum number of tokens is 1000.

For the prompt evaluator $\mathtt{E}$, we used the common facebook/opt-125m model, a decoder-only pre-trained Transformer language model \cite{susan2022opt}. To adapt this model with preference learning, we changed the last layer to output a scalar score. The training setup is facilitated by the Hugging Face Transformers library. We opt for this model due to its commonly used lightweight design for handling text inputs.

We used the following hyperparameters for training the prompt evaluator $\mathtt{E}$. We used AdamW optimizer with $\beta_1 = 0.9$ and $\beta_2 = 0.95$. The training process uses the weight decay of 0.1 and the batch size of 16. The model was trained in 30 epochs.

\subsection{Baselines}
We compare our APS approach with other strong prompt optimization methods. To ensure fairness, we choose methods that use similar-sized downstream LLM $\mathtt{M}$, in particular those that have hundreds of billions of parameters (e.g., Chat-GPT3). In particular, in \textbf{No Prompt}, we directly ask the LLM $\mathtt{M}$ to answer the question with only provided context but not any instructions or prompts. This is the simplest form of the question-answering task that helps us to prove the importance of a prompt to an LLM in later baselines.
\textbf{Fixed Prompt} is the basic setting of our proposed method. It uses one prompt for all samples in test sets. We chose this single prompt as the one with the highest accuracy tested on training sets. This baseline is to verify the necessity of per-sample prompts if they surpass this single fixed prompt. \textbf{Naive Prompt} represents a naive approach in prompt generation where we simply ask Chat-GPT3.5 to generate the prompt for the current input. 
\textbf{Auto CoT} \cite{zhang2022automatic} is an upgraded chain-of-thought prompting technique employing clustering to build demonstration examples.
\textbf{OPRO} (Optimization by PROmpting) \cite{yang2023large} is a strong method for prompt creation. It collects the past generated prompts and their accuracy as history and asks the generator LLM $\mathtt{A}_1$ to generate a new prompt with a higher accuracy. At testing time, the prompt with the highest score in the history in the training phase is selected and applied to all testing samples. This is an advanced version of the fixed prompt where we conditionally generate new prompts with probably higher accuracies other than the unconditional generation in the previous baseline. \textbf{APE} \cite{zhou2022large} leverages input-output data to instruct the LLM $\mathtt{A}_1$ to generate prompts and assess them based on the downstream LLM's performance. The found prompt is then used for all inputs during testing time. This baseline is closely related to our approach and can considered as part of our APS. To make the baseline more robust, we create another variant of APE named \textbf{APE (voting)} by asking APE to find 5 prompts to solve the task and do voting similarly to Eq. \eqref{eq:ranking}. 


\begin{table}[!t]
\setlength{\tabcolsep}{2.5pt}
\centering
\caption{APS with and without voting.}\label{tab:ensemble-non-ensemble}
\begin{tabular}{l c c c} 
 \hline
  Method     & GSM8K & MultiArith & AQuA \\ \hline
 APS (no voting) & 79.14 & 99.45 & 63.78 \\
APS ( top-3 voting) & 80.88 & 99.45 & 61.81 \\
 APS (top-5 voting) & \textbf{81.49} & \textbf{100} & \textbf{64.57} \\
 APS (top-10 voting) & 81.49 & 100 & 61.81 \\
 \hline
\end{tabular}
\end{table}
\begin{table}[t]
\centering
\setlength{\tabcolsep}{4.5pt}
\caption{APS with and without clustering in both voting and no voting settings.}\label{tab:clustering}
\begin{tabular}{l c c c c} 
 \hline
  Cluster & Vote & GSM8K & MultiArith & AQuA \\ \hline
  \XSolidBrush & \XSolidBrush & 75.64 & 97.70 & 59.45 \\
    \XSolidBrush & \CheckmarkBold & 78.53 & 99.45 &  58.66\\ \hline
    \CheckmarkBold & \XSolidBrush & 79.14 & 99.45 & 63.78 \\
    \CheckmarkBold & \CheckmarkBold & \textbf{81.49} & \textbf{100} & \textbf{64.57} \\ \hline
\end{tabular}
\end{table}

\begin{table}[!t]
\centering
\caption{Hyperparameter sensitivity test.}\label{tab:clustering-more-experiments}
\begin{tabular}{c c c c} 
 \hline
  $c$/$n_p$ & GSM8K & MultiArith & AQuA \\ \hline
  10/3 & 81.49 & 100 & 64.57 \\
  10/5 & 81.49 & 100 & 64.57 \\ \hline
  20/3 &  81.56 & 100 & 64.96 \\
  20/5 & \textbf{81.63} & \textbf{100} & \textbf{65.35} \\
 \hline
\end{tabular}
\end{table}
\begin{table*}[t]
\centering
\small
\caption{Case study from AQUA dataset: Prompts that have higher scores and from the same cluster as the input tend to be better and achieve better accuracy. The actual answers are much longer. We only show the extracted final output (bold).}\label{tab:output}
\begin{tabular}{p{12cm}cc} \hline
\multicolumn{3}{l}{\begin{tabular}[c]{@{}l@{}}\textbf{Question}: Of the 200 students in a school, at least 45\% attended the prom night and at least 35\% took part in the debating session. \\What is the maximum number of students who could have neither attended the prom night nor the debating session? \\
Options: ['A)27', 'B)81', 'C)90', 'D)99', 'E)110']\end{tabular}} \\ 
Correct answer: E \\ \hline

Prompt       & Answer & Score      \\ \hline

\begin{tabular}[c]{@{}l@{}}\textbf{\textcolor{blue}{Good prompt (same cluster)}}: 
Given a mathematical question, the AI should follow these steps to solve it: \\
1. Read and understand the question thoroughly. \\
2. Identify the relevant mathematical concepts required to solve the problem. \\
3. Organize the given information and identify the unknown variable. \\
4. Use the relevant mathematical formulae to form an equation or system of equations. \\
5. Solve the equation(s) to obtain the value of the unknown variable. \\
6. Verify the solution by checking if it satisfies the given conditions. \\
7. If the solution is not correct, recheck the steps and calculations and identify any errors. \\ 
8. If no errors are found, check for any omitted or extraneous solutions. \\ 
9. Present the answer in the required format, such as a fraction or decimal. \\
10. Double-check the answer for any errors or mistakes. \\

The AI should follow these steps and provide the answer to the given mathematical problem. \\\end{tabular} 
& 







... \textbf{E)110}  & 0.98  \\ \hline

\begin{tabular}[c]{@{}l@{}}\textbf{\textcolor{red}{Bad prompt (different cluster)}}: Imagine you are a math teacher and the AI is your student. \\ Your task is to guide the AI to solve the given mathematical problem. \\ Begin by highlighting the variables in the problem and their relationships.\\ Then, define the formula or equation that suits the problem\\ and apply it to solve for the unknown variable.\\ Finally, double-check the solution by verifying it with the provided options. \\Remember, the AI has basic knowledge of mathematical \\operations, so explain every step in plain English. \\Output the complete guideline for the AI to solve the mathematical problem.
\\\end{tabular}   
& 

... \textbf{None}.     
& 0.05        \\ \hline




\end{tabular}
\end{table*}


\subsection{Key Hyperparameters}
There are several hyperparameters specific to our methods. Specifically, throughout our experiments, we used a fixed number of clusters ($c = 10$), prompts per cluster ($n_p = 3$), number of demonstrations per meta-prompt ($|\mathcal{D}_{demo}|=10$), number of examples per cluster in synthesizing data ($m=10$), number of top-$k$ voting ($k=5$). We determined that these values strike the optimal balance between performance and training cost. As a result, for each dataset, we had 30 prompts in the prompt database and 3000 synthetic data samples for training the prompt evaluator. A detailed ablation study and hyperparameter analysis is shown in Section \ref{subsec:abl}.

\subsection{The cost of training APS}
Training APS requires $\mathcal{O}(|\mathcal{P}| + c \times m \times |\mathcal{P}|)$ API call to GPT-3.5, where:
$|\mathcal{P}|$ is the prompt database size, $c$ is the number of clusters and $m$ is the number of examples per cluster that are closest to the cluster center.

 In total, prompt and synthetic data generation steps cost between $\$5-10$ for 30 prompts in the prompt database and 3000 synthetic data samples. The total financial cost for all the experiments is $\approx \$30$. Also, it costs $\approx \$10$ for evaluation on all the test sets.
\begin{table}[!t]
\setlength{\tabcolsep}{3.5pt}
\centering
\caption{QA accuracy on math reasoning datasets. $\dag$ denotes results reported in the original papers while $\diamondsuit$ indicates results run by our machines using the same setup as our method.}\label{tab:main-results}

\begin{tabular}{l c c c} 
 \hline
  Method     & GSM8K & MultiArith & AQuA \\ \hline
 No Prompt$^\diamondsuit$ & 74.60 & 98.89 & 58.26 \\ 
 Fixed Prompt$^\diamondsuit$ & 78.45 & 97.22 & 59.45 \\
 Naive Prompt$^\diamondsuit$ & 58.76 & 90.00 & 56.59 \\
 Auto CoT$^\dag$ & 62.8 & 93.20 & 36.50 \\
  OPRO$^\dag$ & 80.20 & 95.30 & 54.30 \\
 APE$^\diamondsuit$ & 76.02 & 98.33 & 61.81 \\
 APE (voting)$^\diamondsuit$ & 77.61 & 99.45 & 61.14 \\

 \hline
 APS (Our) & \textbf{81.49} & \textbf{100} & \textbf{64.57} \\
 \hline
\end{tabular}

\end{table}

\section{Results and Discussion}

In this section, we first present a comparison between our method APS and other baselines. We then conduct an ablation study to assess the contribution of each component and the impact of hyperparameters in the model. Finally, we offer several case studies illustrating the behavior of APS.

\subsection{Effectiveness of Automatic Prompt Selection (APS)}

Table \ref{tab:main-results} summarizes the performance comparison in terms of accuracy. The proposed APS method surpasses all other baselines with a significant gap.
APS performs better of $1.28\%$ than the second-best baseline on GSM8K, $0.55\%$ on MultiArith, and $2.76\%$ on AQuA. To our knowledge, AQuA is the most difficult benchmark with more complex questions when most of them are taken from GMAT (Graduate Management Admission Test) and GRE (General Test) compared to questions using basic arithmetic operation questions $\pm\times\div$ like MultiArith and GSM8K data. Therefore, the optimized prompts contribute a more significant role with a  bigger gap on this dataset.
Compared to the stronger APE with voting, our method shows an improvement of approximately $4\%$ on GSM8K, $0.5\%$ on MultiArith, and $3\%$ on AquaA thanks to our additional components like cluster-based generation and the prompt evaluator.
Compared to another LLM-based prompting approach called OPRO, our APS creates even a larger gap on MultiArith and AQuA due to the per-sample advantage of leveraging a prompt evaluator.
On the other hand, while AutoCoT and Naive Prompt generate prompts for each sample, these methods perform significantly behind other baselines, including No Prompt. We hypothesize that without appropriate optimization, per-sample prompting is unstable and may impede downstream performance. This underscores the significance of our approach in introducing a stable procedure through the use of a prompt evaluator.




\subsection{Ablation Study}\label{subsec:abl}
In this section, we investigate the role of the voting mechanism and the importance of the clustering step in our approach. We also examine the performance of our method when varying important hyperparameters such as the number of clusters $c$ and the number of prompts per cluster $n_p$.

\paragraph{The role of voting}

As mentioned in Section \ref{sec:ranking}, the prompt evaluator allows us to extract the list of prompts with the highest scores for a given input question-context pair. In this experiment, we investigate the actual performance gap between using a voting mechanism and not using it. According to Table \ref{tab:ensemble-non-ensemble}, the best setting top-5 voting shows consistent improvement of approximately $2\%$ on GSM8K against no-voting, $0.5\%$ on MultiArith, and $1\%$ on AQuA. Interestingly, the choice of voting numbers has to be selected carefully to lead to the best performance. Specifically, in a challenging dataset like AQuA, both the selection of top-3 and top-10 lead to the same suboptimal performance even lower than the no-voting setting. Using too few votes limits the potential diversity of the voting models while using too many votes can introduce noises from other less certain predictions.



\paragraph{The importance of generating the cluster-based prompt database}
According to Table \ref{tab:clustering}, the generated cluster-based prompt database always performs better than not using clustering information. Specifically, when not using voting, the performance of the cluster-based approach is lower than that of using voting as nearly $4\%$ gap on GSM8K, $2\%$ on MultiArith, and $5\%$ on AQuA. Similarly, when both approaches use a voting mechanism, the cluster-based approach still performs better than the non-cluster one around $3\%$ on GSM8K, $0.5\%$ on MultiArith, and $5\%$ on AQuA. As presented in Section \ref{subsec:prompt_database}, similar inputs should share the same prompt, and different inputs should be paired with different prompts. Therefore, our proposed cluster-based prompt generation plays a very important role in the whole APS architecture.

\paragraph{Hyperparameter sensitivity of $c$ and $n_p$}
We tried different combinations of the number of clusters $c$ and the number of prompts $n_p$ and report the results in Table \ref{tab:clustering-more-experiments}.
Overall, the results show that increasing the number of clusters and generating more prompts per cluster are beneficial for improving performance. We hypothesize that this enables a more diverse prompt database and thus increases the robustness of the approach. However, due to the significant high cost of increasing $c$ and $n_p$ in exchange for a small gain, we selected $c=10$ and $n_p=3$ across all other experiments.

\subsection{Case Study}

In this section, we analyze the quality of our prompt generation for one question from AQUA dataset (see Table \ref{tab:output}). The question involves complex algebra, necessitating multiple reasoning steps, unlike general questions answerable by LLMs using "role-playing" prompts. Our prompt evaluator selects a prompt with the highest score of $0.98$ (a good prompt) for this specific question. As anticipated, the chosen prompt aligns with the cluster associated with the input question, offering a detailed and sophisticated step-by-step guideline for the LLM to produce the correct answer. For comparison, we choose a low-score (bad) prompt with a score of $0.05$, which is a "role-playing" type prompt belonging to a different cluster. Consequently, it is unsuitable for the question, leading to an incorrect answer.




\section{Conclusion}
In this paper, we have presented Automatic Prompt Selection (APS), a novel method to automate prompting for Large Language Models (LLMs). Our method marries prompt generation and prompt ranking. It reduces the prompt search space by grouping similar inputs into clusters and generating prompts for each cluster. It then trains a prompt evaluator to score the prompts based on their ability to guide the LLM to produce outputs that match the correct answers.
Experimental results on three benchmark QA datasets show two important points.
First, the proposed method can select appropriate prompts for different inputs.
Second, the method achieves competitive performance in the zero-shot setting of the QA task on GSM8K, MultiArith, and AQuA datasets.
As future work, we plan to extend our method to few-shot in-context learning and apply it to various NLP tasks.
\cleardoublepage

\section*{Limitations}

Given the automated generation of the prompt database, ensuring the quality of candidate prompts can be tricky. Notably, issues such as prompt duplication may arise despite our efforts to eliminate exact matches. Semantically equivalent prompts with minor token variations may persist, diminishing both the prompt database's diversity and our approach's flexibility. To mitigate this, a potential solution involves assessing the semantic similarity between prompts and retaining only those that exhibit distinctiveness. In addition, implementing our approach involves several steps and introduces additional hyperparameters, a process that may be time-consuming during tuning. While we have diligently tuned these parameters and verified their effectiveness on the datasets under examination, we recommend practitioners explore varying hyperparameter values to optimize results for their specific tasks.

\section*{Ethics Statement}
All datasets and baseline models experimented in this work have no unethical applications or risky broader impacts.
The evaluation uses public datasets that are widely used for mathematical QA reasoning.
It does not include any confidential or personal information of workers or companies.
The baseline methods used for evaluation can be publicly accessed with GitHub links. There is no bias for the re-implementation that can affect the final results.
 The primary objective of our work is to automate the prompt engineering task for Large Language Models (LLMs), aspiring to enhance the practical application of LLMs. Our intentions are sincere, and we do not foresee immediate adverse consequences. Nevertheless, we acknowledge the potential pitfalls should our method be employed to augment large language models, leading to the generation of hallucinated or negative content. Such issues are common in prompt generation approaches, and we are committed to proactively addressing and preventing these concerns to the best of our ability.

\bibliography{custom}

\begin{thebibliography}{30}
\expandafter\ifx\csname natexlab\endcsname\relax\def\natexlab#1{#1}\fi

\bibitem[{Bradley and Terry(1952)}]{bradley1952rank}
Ralph~Allan Bradley and Milton~E Terry. 1952.
\newblock Rank analysis of incomplete block designs: I. the method of paired
  comparisons.
\newblock \emph{Biometrika}, 39(3/4):324--345.

\bibitem[{Brown et~al.(2020)Brown, Mann, Ryder, Subbiah, Kaplan, Dhariwal,
  Neelakantan, Shyam, Sastry, Askell et~al.}]{brown2020language}
Tom Brown, Benjamin Mann, Nick Ryder, Melanie Subbiah, Jared~D Kaplan, Prafulla
  Dhariwal, Arvind Neelakantan, Pranav Shyam, Girish Sastry, Amanda Askell,
  et~al. 2020.
\newblock Language models are few-shot learners.
\newblock \emph{Advances in neural information processing systems},
  33:1877--1901.

\bibitem[{Christiano et~al.(2017)Christiano, Leike, Brown, Martic, Legg, and
  Amodei}]{christiano2017deep}
Paul~F Christiano, Jan Leike, Tom Brown, Miljan Martic, Shane Legg, and Dario
  Amodei. 2017.
\newblock Deep reinforcement learning from human preferences.
\newblock \emph{Advances in neural information processing systems}, 30.

\bibitem[{Deng et~al.(2022)Deng, Wang, Hsieh, Wang, Guo, Shu, Song, Xing, and
  Hu}]{deng2022rlprompt}
Mingkai Deng, Jianyu Wang, Cheng-Ping Hsieh, Yihan Wang, Han Guo, Tianmin Shu,
  Meng Song, Eric Xing, and Zhiting Hu. 2022.
\newblock Rlprompt: Optimizing discrete text prompts with reinforcement
  learning.
\newblock In \emph{Proceedings of the 2022 Conference on Empirical Methods in
  Natural Language Processing}, pages 3369--3391.

\bibitem[{Gao et~al.(2021)Gao, Fisch, and Chen}]{gao2021making}
Tianyu Gao, Adam Fisch, and Danqi Chen. 2021.
\newblock Making pre-trained language models better few-shot learners.
\newblock In \emph{Proceedings of the 59th Annual Meeting of the Association
  for Computational Linguistics and the 11th International Joint Conference on
  Natural Language Processing (Volume 1: Long Papers)}, pages 3816--3830.

\bibitem[{Karl~Cobbe et~al.(2021)}]{karl2021gsm8k}
Mohammad Bavarian Mark Chen Heewoo Jun Lukasz Kaiser Matthias Plappert Jerry
  Tworek Jacob Hilton Reiichiro~Nakano Karl~Cobbe, Vineet~Kosaraju et~al. 2021.
\newblock Training verifiers to solve math word problems.
\newblock \emph{arXiv preprint arXiv:2110.14168}.

\bibitem[{Liao et~al.(2022)Liao, Zheng, and Yang}]{liao2022zero}
Chonghua Liao, Yanan Zheng, and Zhilin Yang. 2022.
\newblock Zero-label prompt selection.
\newblock \emph{arXiv preprint arXiv:2211.04668}.

\bibitem[{Lu et~al.(2022)Lu, Bartolo, Moore, Riedel, and
  Stenetorp}]{lu2022fantastically}
Yao Lu, Max Bartolo, Alastair Moore, Sebastian Riedel, and Pontus Stenetorp.
  2022.
\newblock Fantastically ordered prompts and where to find them: Overcoming
  few-shot prompt order sensitivity.
\newblock In \emph{Proceedings of the 60th Annual Meeting of the Association
  for Computational Linguistics (Volume 1: Long Papers)}, pages 8086--8098.

\bibitem[{Lyu et~al.(2023)Lyu, Jiang, Zeng, Xia, and Luo}]{lyu2023llm}
Hanjia Lyu, Song Jiang, Hanqing Zeng, Yinglong Xia, and Jiebo Luo. 2023.
\newblock Llm-rec: Personalized recommendation via prompting large language
  models.
\newblock \emph{arXiv preprint arXiv:2307.15780}.

\bibitem[{MacQueen et~al.(1967)}]{macqueen1967some}
James MacQueen et~al. 1967.
\newblock Some methods for classification and analysis of multivariate
  observations.
\newblock In \emph{Proceedings of the fifth Berkeley symposium on mathematical
  statistics and probability}, volume~1, pages 281--297. Oakland, CA, USA.

\bibitem[{Nguyen et~al.(2023)Nguyen, Nguyen, Sabahi, Le, Yang, and
  Hotta}]{nguyen2023giant}
Minh-Tien Nguyen, Duy-Hung Nguyen, Shahab Sabahi, Hung Le, Jeff Yang, and
  Hajime Hotta. 2023.
\newblock When giant language brains just aren't enough! domain pizzazz with
  knowledge sparkle dust.
\newblock \emph{arXiv preprint arXiv:2305.07230}.

\bibitem[{Pang et~al.(2023)Pang, Yavuz, Xiong, and
  Zhou}]{pang-etal-2023-sharpt}
Bo~Pang, Semih Yavuz, Caiming Xiong, and Yingbo Zhou. 2023.
\newblock \href {https://doi.org/10.18653/v1/2023.findings-eacl.92}
  {{S}har{PT}: Shared latent space prompt tuning}.
\newblock In \emph{Findings of the Association for Computational Linguistics:
  EACL 2023}, pages 1244--1250, Dubrovnik, Croatia. Association for
  Computational Linguistics.

\bibitem[{Prasad et~al.(2023)Prasad, Hase, Zhou, and Bansal}]{prasad2023grips}
Archiki Prasad, Peter Hase, Xiang Zhou, and Mohit Bansal. 2023.
\newblock Grips: Gradient-free, edit-based instruction search for prompting
  large language models.
\newblock In \emph{Proceedings of the 17th Conference of the European Chapter
  of the Association for Computational Linguistics}, pages 3827--3846.

\bibitem[{Qin and Eisner(2021)}]{qin2021learning}
Guanghui Qin and Jason Eisner. 2021.
\newblock Learning how to ask: Querying lms with mixtures of soft prompts.
\newblock In \emph{Proceedings of the 2021 Conference of the North American
  Chapter of the Association for Computational Linguistics: Human Language
  Technologies}, pages 5203--5212.

\bibitem[{Reimers and Gurevych(2019)}]{reimers-2019-sentence-bert}
Nils Reimers and Iryna Gurevych. 2019.
\newblock \href {https://arxiv.org/abs/1908.10084} {Sentence-bert: Sentence
  embeddings using siamese bert-networks}.
\newblock In \emph{Proceedings of the 2019 Conference on Empirical Methods in
  Natural Language Processing}. Association for Computational Linguistics.

\bibitem[{Reynolds and McDonell(2021)}]{reynolds2021prompt}
Laria Reynolds and Kyle McDonell. 2021.
\newblock Prompt programming for large language models: Beyond the few-shot
  paradigm.
\newblock In \emph{Extended Abstracts of the 2021 CHI Conference on Human
  Factors in Computing Systems}, pages 1--7.

\bibitem[{Rik Koncel-Kedziorski and Hajishirzi(2016)}]{rik2016mathworld}
Aida Amini Nate~Kushman Rik Koncel-Kedziorski, Subhro~Roy and Hannaneh
  Hajishirzi. 2016.
\newblock Mawps: A math word problem repository.
\newblock In \emph{Proceedings of the 2016 conference of the north american
  chapter of the association for computational linguistics: human language
  technologies)}, pages 1152–--1157.

\bibitem[{Roy and Roth(2016)}]{roy2016multiarith}
Subhro Roy and Dan Roth. 2016.
\newblock Solving general arithmetic word problems.
\newblock \emph{arXiv preprint arXiv:1608.01413}.

\bibitem[{Singh et~al.(2022)Singh, Morris, Aneja, Rush, and
  Gao}]{singh2022explaining}
Chandan Singh, John~X Morris, Jyoti Aneja, Alexander~M Rush, and Jianfeng Gao.
  2022.
\newblock Explaining patterns in data with language models via interpretable
  autoprompting.
\newblock \emph{arXiv preprint arXiv:2210.01848}.

\bibitem[{Sorensen et~al.(2022)Sorensen, Robinson, Rytting, Shaw, Rogers,
  Delorey, Khalil, Fulda, and Wingate}]{sorensen2022information}
Taylor Sorensen, Joshua Robinson, Christopher Rytting, Alexander Shaw, Kyle
  Rogers, Alexia Delorey, Mahmoud Khalil, Nancy Fulda, and David Wingate. 2022.
\newblock An information-theoretic approach to prompt engineering without
  ground truth labels.
\newblock In \emph{Proceedings of the 60th Annual Meeting of the Association
  for Computational Linguistics (Volume 1: Long Papers)}, pages 819--862.

\bibitem[{Srivastava et~al.(2023)Srivastava, Rastogi, Rao, Shoeb, Abid, Fisch,
  Brown, Santoro, Gupta, Garriga-Alonso et~al.}]{srivastava2023beyond}
Aarohi Srivastava, Abhinav Rastogi, Abhishek Rao, Abu Awal~Md Shoeb, Abubakar
  Abid, Adam Fisch, Adam~R Brown, Adam Santoro, Aditya Gupta, Adri{\`a}
  Garriga-Alonso, et~al. 2023.
\newblock Beyond the imitation game: Quantifying and extrapolating the
  capabilities of language models.
\newblock \emph{Transactions on Machine Learning Research}.

\bibitem[{Susan~Zhang(2022)}]{susan2022opt}
Naman Goyal Mikel Artetxe Moya Chen Shuohui Chen Christopher Dewan Mona Diab
  Xian Li Xi Victoria Lin Todor Mihaylov Myle Ott Sam Shleifer Kurt Shuster
  Daniel Simig Punit Singh Koura Anjali Sridhar Tianlu Wang Luke~Zettlemoyer
  Susan~Zhang, Stephen~Roller. 2022.
\newblock Opt: Open pre-trained transformer language models.
\newblock \emph{arXiv preprint arXiv:1608.01413}.

\bibitem[{Wang~Ling and Blunsom(2017)}]{ling2017aqua}
Chris~Dyer Wang~Ling, Dani~Yogatama and Phil Blunsom. 2017.
\newblock Program induction by rationale generation: Learning to solve and
  explain algebraic word problems.
\newblock \emph{arXiv preprint arXiv:1705.04146}.

\bibitem[{Wei et~al.(2022)Wei, Wang, Schuurmans, Bosma, Xia, Chi, Le, Zhou
  et~al.}]{wei2022chain}
Jason Wei, Xuezhi Wang, Dale Schuurmans, Maarten Bosma, Fei Xia, Ed~Chi, Quoc~V
  Le, Denny Zhou, et~al. 2022.
\newblock Chain-of-thought prompting elicits reasoning in large language
  models.
\newblock \emph{Advances in Neural Information Processing Systems},
  35:24824--24837.

\bibitem[{White et~al.(2023)White, Fu, Hays, Sandborn, Olea, Gilbert, Elnashar,
  Spencer-Smith, and Schmidt}]{white2023prompt}
Jules White, Quchen Fu, Sam Hays, Michael Sandborn, Carlos Olea, Henry Gilbert,
  Ashraf Elnashar, Jesse Spencer-Smith, and Douglas~C Schmidt. 2023.
\newblock A prompt pattern catalog to enhance prompt engineering with chatgpt.
\newblock \emph{arXiv preprint arXiv:2302.11382}.

\bibitem[{Yang et~al.(2023)Yang, Wang, Lu, Liu, Le, Zhou, and
  Chen}]{yang2023large}
Chengrun Yang, Xuezhi Wang, Yifeng Lu, Hanxiao Liu, Quoc~V Le, Denny Zhou, and
  Xinyun Chen. 2023.
\newblock Large language models as optimizers.
\newblock \emph{arXiv preprint arXiv:2309.03409}.

\bibitem[{Yao et~al.(2022)Yao, Zhao, Yu, Du, Shafran, Narasimhan, and
  Cao}]{yao2022react}
Shunyu Yao, Jeffrey Zhao, Dian Yu, Nan Du, Izhak Shafran, Karthik~R Narasimhan,
  and Yuan Cao. 2022.
\newblock React: Synergizing reasoning and acting in language models.
\newblock In \emph{The Eleventh International Conference on Learning
  Representations}.

\bibitem[{Zhang et~al.(2022{\natexlab{a}})Zhang, Wang, Zhou, Schuurmans, and
  Gonzalez}]{zhang2022tempera}
Tianjun Zhang, Xuezhi Wang, Denny Zhou, Dale Schuurmans, and Joseph~E Gonzalez.
  2022{\natexlab{a}}.
\newblock Tempera: Test-time prompt editing via reinforcement learning.
\newblock In \emph{The Eleventh International Conference on Learning
  Representations}.

\bibitem[{Zhang et~al.(2022{\natexlab{b}})Zhang, Zhang, Li, and
  Smola}]{zhang2022automatic}
Zhuosheng Zhang, Aston Zhang, Mu~Li, and Alex Smola. 2022{\natexlab{b}}.
\newblock Automatic chain of thought prompting in large language models.
\newblock In \emph{The Eleventh International Conference on Learning
  Representations}.

\bibitem[{Zhou et~al.(2022)Zhou, Muresanu, Han, Paster, Pitis, Chan, and
  Ba}]{zhou2022large}
Yongchao Zhou, Andrei~Ioan Muresanu, Ziwen Han, Keiran Paster, Silviu Pitis,
  Harris Chan, and Jimmy Ba. 2022.
\newblock Large language models are human-level prompt engineers.
\newblock In \emph{The Eleventh International Conference on Learning
  Representations}.

\end{thebibliography}
\bibliographystyle{plain}


\end{document}